\documentclass[letterpaper]{article} 
\usepackage{aaai24}  
\usepackage{times}  
\usepackage{helvet}  
\usepackage{courier}  
\usepackage[hyphens]{url}  
\usepackage{graphicx} 
\usepackage{natbib}  
\usepackage{caption} 
\usepackage{algorithm}
\usepackage{algorithmic}

\usepackage{newfloat}
\usepackage{listings}
\DeclareCaptionStyle{ruled}{labelfont=normalfont,labelsep=colon,strut=off} 
\floatstyle{ruled}
\newfloat{listing}{tb}{lst}{}
\floatname{listing}{Listing}
\usepackage{color}
\usepackage{mathtools}
\usepackage{amsfonts,amssymb}
\usepackage{soul}

\usepackage{subfig}
\usepackage{adjustbox}

\usepackage{array,xcolor,colortbl}
\usepackage{multirow}

\usepackage[capitalize]{cleveref}
\crefname{section}{Sec.}{Secs.}
\Crefname{section}{Section}{Sections}
\Crefname{table}{Table}{Tables}
\crefname{table}{Tab.}{Tabs.}
\def\eg{e.g.} 
\def\ie{i.e}

\newcommand{\Tref}[1]{Table~\ref{#1}}

\newcommand{\Eref}[1]{Eq.~(\ref{#1})}
\newcommand{\fref}[1]{Fig.~\ref{#1}}

\newcommand{\sota}{state-of-the-art}

\newcommand{\cP}{\mathcal{P}}
\newcommand{\cF}{\mathcal{F}}
\newcommand{\cS}{\mathcal{S}}
\newcommand{\cL}{\mathcal{L}}

\newcommand{\bu}{\mathbf{u}}
\newcommand{\br}{\mathbf{r}}
\newcommand{\bc}{\mathbf{c}}
\newcommand{\bC}{\mathbf{C}}
\newcommand{\bl}{l}
\newcommand{\bw}{\mathbf{w}}
\newcommand{\bL}{\mathbf{L}}
\newcommand{\bY}{\mathbf{Y}}

\newcommand{\bZ}{\mathbf{Z}}
\newcommand{\bz}{\mathbf{z}}

\newcommand{\CT}{CT$^2$}

\title{Colorizing Monochromatic Radiance Fields}

\author{
    Yean Cheng\textsuperscript{\rm 1,\rm 2,\rm3},
    Renjie Wan\textsuperscript{\rm 4\footnote{Corresponding authors.}},
    Shuchen Weng\textsuperscript{\rm 1,\rm 2},
    Chengxuan Zhu\textsuperscript{\rm 5},\\
    Yakun Chang\textsuperscript{\rm 1,\rm 2},
    Boxin Shi\textsuperscript{\rm 1,\rm 2,\rm 3\footnotemark[1]}
}
\affiliations{
    \textsuperscript{\rm 1}National Key Laboratory for Multimedia Information Processing, School of Computer Science, Peking University\\
    \textsuperscript{\rm 2}National Engineering Research Center of Visual Technology, School of Computer Science, Peking University\\
    \textsuperscript{\rm 3}AI Innovation Center, School of Computer Science, Peking University\\
    \textsuperscript{\rm 4}Department of Computer Science, Hong Kong Baptist University\\
    \textsuperscript{\rm 5}National Key Lab of General AI, School of Intelligence Science and Technology, Peking University\\
    cya17@stu.pku.edu.cn, renjiewan@hkbu.edu.hk, \{shuchenweng, peterzhu, yakunchang, shiboxin\}@pku.edu.cn

}

\begin{document}

\maketitle

\begin{abstract}

Though Neural Radiance Fields (NeRF) can produce colorful 3D representations of the world by using a set of 2D images, such ability becomes non-existent when only monochromatic images are provided. Since color is necessary in representing the world, reproducing color from monochromatic radiance fields becomes crucial. To achieve this goal, instead of manipulating the monochromatic radiance fields directly, we consider it as a representation-prediction task in the \textit{Lab} color space. By first constructing the luminance and density representation using monochromatic images, our prediction stage can recreate color representation on the basis of an image colorization module. We then reproduce a colorful implicit model through the representation of luminance, density, and color. Extensive experiments have been conducted to validate the effectiveness of our approaches. Our project page: https://liquidammonia.github.io/color-nerf.
\end{abstract}

\section{Introduction}

\begin{figure}[t]
    \centering
        \includegraphics[width=\linewidth]{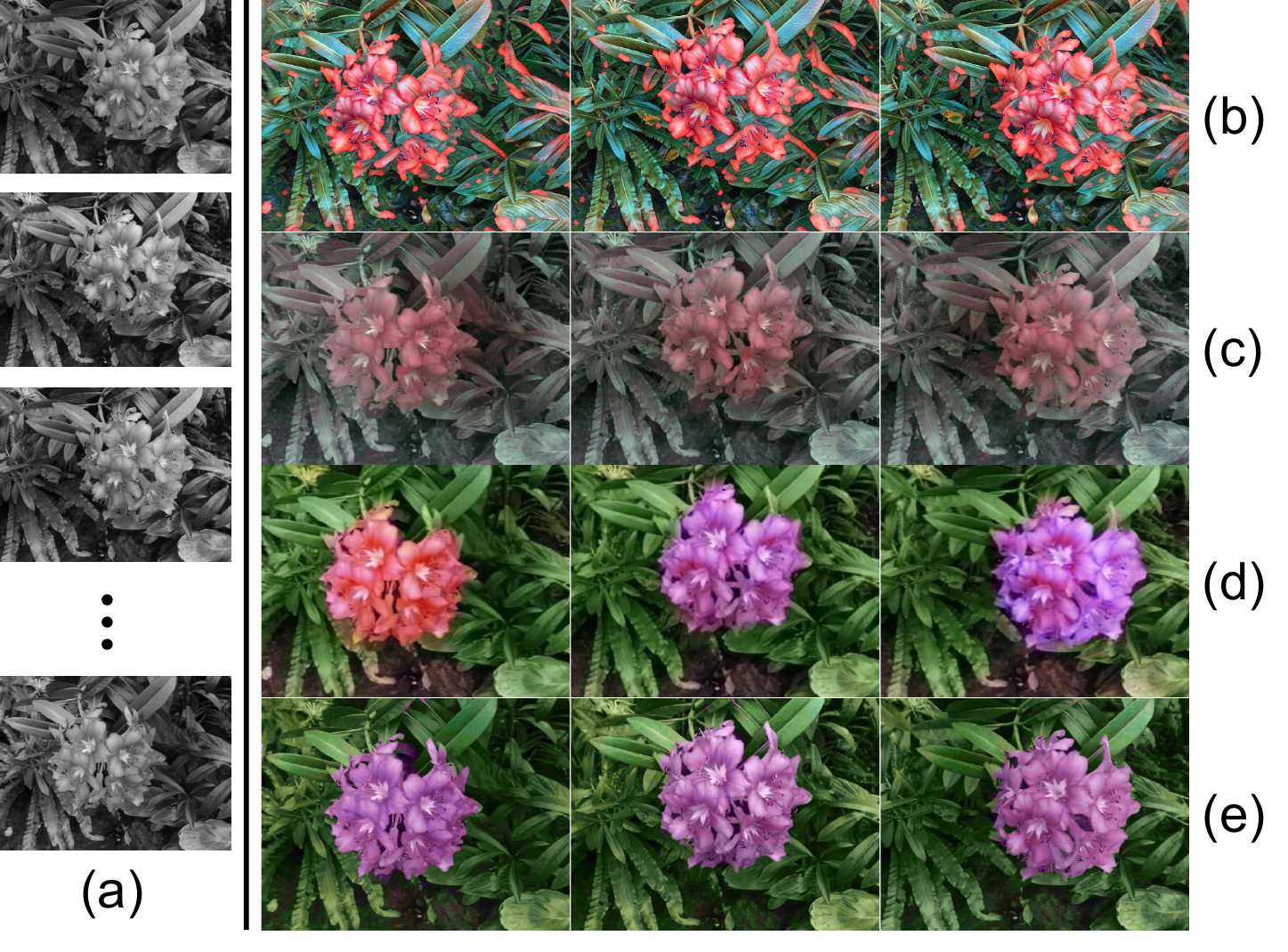}
    \caption{With multi-view monochromatic images (a) as inputs, (b-e) are three novel views synthesised by NeRF models. Existing color editing NeRFs (b) ARF~\cite{zhang2022arf} and (c) CLIP-NeRF~\cite{wang2022clipnerf} could not guarantee pixel-wise color adherence, while using ``colorize-then-fuse" solution (d) \CT~\cite{WengCT2}+NeRF) suffers from color inconsistency across different views. The proposed ColorNeRF (e) can generate a more plausible and vivid colorized NeRF compared to previous models. }

    \label{fig:teaser}
\end{figure}

Neural Radiance Fields (NeRF)~\cite{MildenhallSTBRN20} is able to create a colorful 3D representation of the world by using a set of 2D images. \textit{Can this created implicit 3D model still be colorful when only monochromatic images are available?}

The answer is frustrating. The original design of NeRF is unable to create a colorful appearance from monochrome, and colorizing the monochromatic radiance fields using external forces seems to be the only option. Colorization is a classical problem being studied for more than a decade~\cite{ DBLP:conf/iccv/ChengYS15,ironi2005colorization, levin2004colorization,luan2007natural}, with various applications in artistic creation, and legacy photo restoration. During its evolution on images/videos, there are two common standards that a good colorization scheme should follow: 1) \textbf{plausibility}, which requires the colorized results to demonstrate visually reasonable appearance~\cite{iizuka2016let, larsson2016learning}; 2) \textbf{vividness}, which ensures the high level of saturation for colorized results~\cite{WengCT2,wu2021towards, zhang2016colorful, zhang2017realtime}. These standards should be applied to \textbf{colorizing monochromatic radiance fields} as well, but how to achieve this remains an open problem.

Directly manipulating radiance fields seems to be a straightforward way to achieve this goal of colorization. One solution is to regard the color as a kind of ``style" and then transfer the style into radiance fields~\cite{zhang2022arf}. However, as displayed in \fref{fig:teaser}(b), since such a strategy cannot guarantee pixel-wise color adherence, the color can only distribute on radiance fields irregularly, thus violating the plausibility standard. 
A different approach involves manipulating the color attributes in radiance fields directly~\cite{tojo2022posternerf, wang2022clipnerf}. This technique is intended for replacing colors by identifying the current color attributes and replacing them with new ones. However, it is not applicable to monochromatic radiance fields where there are no existing color attributes.
As displayed in \fref{fig:teaser}(c), the inability to perceive color palettes when using ``direct manipulating" approaches~\cite{wang2022clipnerf} on monochromatic radiance fields leaves rendered results below the vividness standard.

Another alternative is the ``colorize-then-fuse" solution, \ie, first colorize monochromatic images and then fuse them for radiance field construction. However, without considering the view-dependent correlation, the examples colorized by image-based approaches cannot guarantee color consistency in the constructed radiance fields, as displayed in \fref{fig:teaser}(d), which also obviously violates the plausibility standard. However, despite the unsatisfactory plausibility across views, this paradigm indeed achieves better vividness than directly manipulating radiance fields, compared with \fref{fig:teaser}(b) and \fref{fig:teaser}(c).
This is partly because of the operation on complementing the missing color channels~\cite{huang2022unicolor, zhang2016colorful, wu2021towards} in the CIE \textit{Lab} color space as a channel-prediction task~\cite{iizuka2016let, WengCT2, zhang2016colorful}, i.e., inferring the missing \textit{a} and \textit{b} channels from the given \textit{L} channel (monochromatic image). As opposed to producing three-channel RGB outputs, such an operation allows the neural network to focus only on the generation of two color channels~\cite{anwar2020image}, which reduces computational costs and uncertainty during colorization.

\begin{figure*}[t]
\centering
\includegraphics[width=1\textwidth]{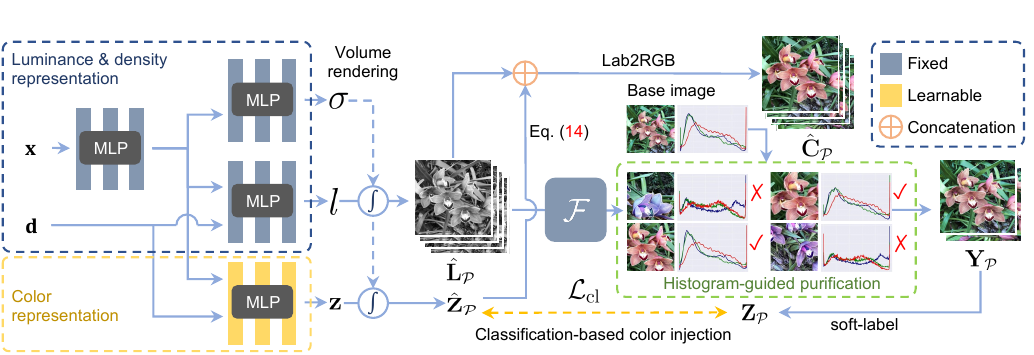}
\caption{\textbf{The overall pipeline of the proposed ColorNeRF}. With rays from multiple viewpoints as inputs, luminance and density representation is first constructed with supervision over the ground truth monochromatic images, yielding monochromatic image patches $\hat{\mathbf{L}}_\cP$. Then we predict color with an off-the-shelf 2D colorization module $\cF(\cdot)$, followed by our histogram-guided purification module to enhance plausibility. Lastly, we inject the color information in $\bZ_\cP$ to the color representation with our classification-based color injection module. The final output $\hat{\bC}_{\mathcal{P}}$ is calculated by the concatenation of $\hat{\bL}_{\mathcal{P}}$ and $\hat{\bY}_{\mathcal{P}}$, followed by the $Lab$ to RGB conversion.
}
\label{fig:pipeline}
\end{figure*}

Based on the above observations, in this paper, we propose \textbf{ColorNeRF}, to colorize monochromatic radiance fields by predicting the missing representation of color channels. As displayed in~\fref{fig:pipeline}, ColorNeRF first builds luminance and density radiance fields by solely using monochromatic images and then infers the color representation for channel $a$ and $b$. Instead of building the radiance fields via directly altering in the image domain like the ``colorize-then-fuse" paradigm, we inject color knowledge into the predicted representations from an off-the-shelf colorization module~\cite{WengCT2} based on a newly proposed query-based colorization strategy. By gradually imposing changes on the predicted representation, our model can finally maintain color consistency for better plausibility. A histogram-guided purification module and a classification-based color injection module are further proposed to better address color plausibility and enhance the vividness. 

To sum up, ColorNeRF is the first approach that achieves rendering plausible and vivid radiance fields from monochromatic images via the following contributions: 
\begin{itemize}
    \item a representation prediction paradigm tailored to recreate colors for monochromatic radiance fields;
    \item a query-based colorization strategy and a histogram-guided purification module for maintaining strong plausibility, and
    \item a classification-based color injection module for achieving high vividness in colorized results.
\end{itemize}

Extensive experiments on LLFF dataset \cite{mildenhall2019llff} and our own captured scenes demonstrate that ColorNeRF achieves \sota{} results with better plausibility and vividness in quantitative and qualitative measurements. We also show the colorful NeRF generated from real monochromatic inputs, \eg,  monochrome photography and classical movies.

\section{Related work} 
\label{subsec: related work: image colorization}

\paragraph{Image colorization.} Several methods have been proposed to address the plausibility and vividness in colorization. Automatic colorization methods use a single monochromatic image as the input. It is a highly ill-posed task while requires estimating two missing color channels from one monochormatic channel. Early approaches rely on the local freature extraction~\cite{guadarrama2017pixcolor, larsson2016learning, zhang2016colorful}. Later, better colorization can be achieved via the generative models~\cite{cao2017unsupervised, vitoria2020chromagan}. Several other studies~\cite{Kim2022BigColor, wu2021towards, zhao2020pixelated} have focused on utilizing external prior knowledge from other low-level vision tasks.
Other works focus on how to inject multi-modal user-guided features to conduct conditional colorization.
For example, stroke-based approaches~\cite{Yun_2023_WACV, zhang2016colorful, zhang2017realtime} and text-based methods~\cite{chang2022lcoder, chen2018language, huang2022unicolor} are proposed to adopt necessary attention features. In terms of image/video-based colorization, the above methods have pushed the boundaries for plausibility and vividness, but still cannot achieve these two goals for monochromatic radiance fields used to implicitly represent 3D space, which we hope to achieve in this paper.

\label{subsec: related work: nerf}
\paragraph{Manipulating colors for NeRF.} NeRF~\cite{MildenhallSTBRN20} is poised to be an effective paradigm to implicitly represent the 3D world. The recent advances show that radiance fields can be robustly constructed when noise~\cite{pearl2022nan}, occlusion~\cite{martinbrualla2020nerfw}, or even blurring phenomena~\cite{ma2022deblur} are encountered. However, the exploration for constructing plausible and vivid radiance fields from monochromatic images is still left unresolved. A number of existing approaches can change the color rendered from NeRF. For example, the approaches~\cite{zhang2022arf, wang2022nerf}) for stylization can conduct the change by transferring the styles from external sources to radiance fields. However, the transferred styles actually distribute in radiance fields irregularly. Recently, several methods are proposed to directly edit the color of radiance fields by extracting the color palette~\cite{tojo2022posternerf, Gong2023RecolorNeRFLD, Kuang2022PaletteNeRFPA} or from external models~\cite{fan2022nerfsos, kobayashi2022distilledfeaturefields, Niemeyer2021GIRAFFE,wang2022clipnerf,liu2021editing}. However, as they rely on established color attributes to distinguish regions to be colorized, they cannot colorize monochromatic radiance fields, which is another issue to be addressed in this paper.

\section{Preliminaries}
\label{subsec: preliminaries}
\label{subsec: 2D colorization}

\paragraph{Colorization for images.} The image colorization task is usually conducted in the CIE \textit{Lab} space~\cite{iizuka2016let,WengCT2, zhang2016colorful} instead of the RGB space, as this color space is device-independent and robust in approximating human vision. The \textit{L} channel represents perceptual lightness and \textit{ab} represents human perceptual colors. When \textit{Lab} color space is employed, the monochromatic image can be considered as an image with a single \textit{L} channel. Thus, the colorization of a monochromatic image can be regarded as transferring the prediction of missing color represented by information in \textit{a} and \textit{b} channels, when only \textit{L} channel is provided, formulated as follows:
\begin{equation}
\label{eq: c=cat LY}
    \bC_{\mathrm{lab}} = \mathrm{concat}\{\bL, \cF(\bL)\},
\end{equation}
where $\cF(\cdot)$ denotes the estimation of \textit{a} and \textit{b} channels and $\bC_{\mathrm{lab}}$ is the estimated results with complete three channels in the $Lab$ color space.

\paragraph{Neural radiance fields.} NeRF~\cite{MildenhallSTBRN20} utilizes multilayer perceptron (MLP) to implicitly represent 3D scene. Taking a point's 3D coordinate $\mathbf{x}\in\mathbb{R}^3$ as input, MLP $\Theta_{\sigma}(\cdot)$ first yields density $\mathbf{\sigma}$ and points encoding $\bw$. MLP $\Theta_{\mathrm{C}}(\cdot, \cdot)$ subsequently takes $\bw$ and view direction $\mathbf{d}\in[-\pi, \pi]^2$ as inputs and predicts $\bc\in\mathbb{R}^3$, denoting the RGB color, summarized as: 
\begin{align}
\label{eq:nerf}
	(\bw, \mathbf{\sigma}) &=\Theta_{\sigma}(\mathbf{x}), \\
	\bc &= \Theta_{\mathrm{C}}(\bw, \mathbf{d}).
\end{align}
In the volume rendering stage, given a camera ray $\br(t) = \mathbf{o} + t\mathbf{d}$, where $t\in[t_{\mathrm{near}}, t_{\mathrm{far}}]$ is the depth, $\mathbf{o}$ is the camera origin; NeRF calculates the perceptual color $\bC(\br)$ of the ray using quadrature of $M$ sampled points:
\begin{equation}
\label{eq:vol rendering}
    \bC(\br) = \sum^M_{m=1}T(m)(1 - \mathrm{exp}(-\sigma_m\delta_m))\bc_m,
\end{equation}
where $T(m) = \mathrm{exp}(-\sum^{m-1}_{l=1}\sigma_l\delta_l)$ and $\delta_m=t_{m+1}-t_m$ are intervals of adjacent sampled points, $(\bc_m, \sigma_m)$ are generated by neural networks. NeRF has a ``one-scene-per-model" property, \ie, a NeRF model is solely optimized on a collection of images and their poses from one scene using photometric re-rendering loss: 
\begin{equation}
\label{eq:photo}
\cL_{\mathrm{photometric}}=\sum_{\br\in \mathcal{R}}||\bC(\br) - \hat{\bC}(\br)||^2_2, 
\end{equation}
where $\mathcal{R}$ is the set of sampled rays, $\bC(\br)$ and $\hat{\bC}(\br)$ are the ground truth and predicted values, respectively.

Given RGB images, NeRF is capable of converting image-level details into a colorful implicit 3D representation by using~\Eref{eq:photo}, while it cannot create color attributes if they are not available from input images. Thus, feeding monochromatic images widely observed in our lives  (\eg, monochrome photography and classical movies) to NeRF can only lead to monochromatic radiance fields. Colorizing such monochromatic radiance fields with high plausibility and vividness is the problem to be solved next.

\section{Proposed method}
\label{sec: method}
The overall pipeline of ColorNeRF is summarized in \fref{fig:pipeline}. According to the analysis in \fref{fig:teaser}, we follow the paradigm established in image colorization~\cite{iizuka2016let,WengCT2, zhang2016colorful} by first constructing a luminance and density representation with monochromatic images and then predicting the missing representation of \textit{a} and \textit{b} channels. After volume rendering, monochromatic image patches are first sent to the colorization module, the results are gradually injected using a query-based colorization strategy, followed by our histogram-guided purification module to remove outliers. The purified image patches are used to supervise the prediction of color representation with a proposed classification-based color injection module.

We utilize patch sampling scheme similar to GRAF~\cite{schwarz2020graf} throughout model training. Specifically, in the training process, the $K\times K$ image patch $\cP(\bu, s)$ is defined by:
\begin{equation}
    \label{eq:patch sampling}
    \cP(\bu, s)\! =\! \left\{ (sx+u, sy+v) | x,y\in \{-\frac{K}{2}, ..., \frac{K}{2} - 1\}\right\},
\end{equation}
where $\bu=(u, v)$ is the center of the image patch, $s$ controls the perception field of the sampled patch. The corresponding 3D rays are determined by $\cP(\bu, s)$. With this sampling scheme, the patches from volume rendering become semantically meaningful, and hence could be fed to an external model for further processing. 

\begin{figure}[t]
    \centering
    \includegraphics[width=\linewidth]{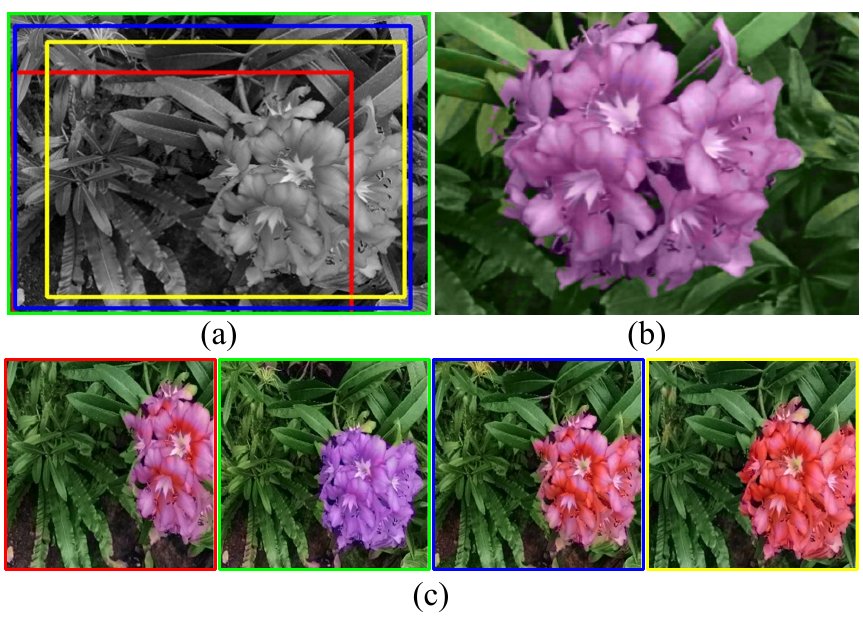}
    \caption{ 
    Our query-based injection strategy could generate plausible results from inconsistent colors. (a) is the input monochromatic image with different sampled patches (denoted as rectangles); (b) is the final output of our model with consistent color; (c) are the results from \CT~\cite{WengCT2}, corresponding to image patches in (a).}
    \label{fig:toy_mean}
\end{figure}

\subsection{Luminance and density representation}
\label{sec:luminancerepr}
We first construct the representation for luminance and density, and then fix them during color representation prediction. The luminance and density representation can be simply defined as:
\begin{equation}
\label{eq:monochromenerf}
    \bl = \Theta_{\mathrm{M}}(\mathbf{w}, \mathbf{d}), 
\end{equation}
where $\bl$ is the rendered monochromatic output, $\Theta_{\mathrm{M}}(\cdot,\cdot)$ is the mapping network in luminance and density representation. COLMAP~\cite{schoenberger2016sfm, schoenberger2016mvs} is used for pose estimation during mapping defined by \Eref{eq:monochromenerf}. Due to that COLMAP's proper functioning requires monochromatic images, our setting does not affect its performance. We first supervise the luminance and density representation using photometric loss similar to \Eref{eq:photo}, 
\begin{equation}
\label{eq:photo_l}
\cL_{\mathrm{photometric}}=\sum_{\mathbf{p} \in \cP}||\bL_\cP(\mathbf{p}) - \hat{\bL}_\cP(\mathbf{p})||^2_2,
\end{equation}
where $\mathbf{p} = p(\mathbf{r})$ is the pixel corresponding to the ray $\mathbf{r}$.

\subsection{Color representation prediction}
\label{sec:colorrepr}
With the luminance and density representation obtained in~\Eref{eq:monochromenerf}, we aim at predicting the color representation via the mapping network $\Theta_{\mathrm{Z}}(\cdot, \cdot)$ below,
\begin{equation}
\label{eq:colornerf}
    \bz = \Theta_{\mathrm{Z}}(\mathbf{w}, \mathbf{d}), 
\end{equation}
where $\bz$ denotes the predicted representation for $ab$ channels. The difficulty encountered by the mapping correlation in~\Eref{eq:colornerf} comes from the lack of color supervision during representation construction. Thus, incorporating color knowledge into predicted representation, while preserving the plausibility and vividness of the results, is crucial.

\paragraph{Query-based colorization. }
Our approach could utilize color information from different off-the-shelf colorization models. Without losing generality, we obtain color knowledge from a \sota{} automatic colorization work \CT~\cite{WengCT2}. For maintaining higher plausibility, rather than colorizing images ahead of representation prediction (\ie, the ``colorize-then-fuse" paradigm), where each image pixel in a camera pose is assigned with a fixed color before optimization, we propose a query-based colorization strategy by first querying the colorization module with the rendered luminance samples and then dynamically injecting color knowledge into the predicted representation. Such a query strategy colorizes each pixel in a camera pose multiple times and incorporates various possible colors into our color representation. As displayed in~\fref{fig:toy_mean}, though the sampled image patches in one image are assigned with different colors during each iteration (\fref{fig:toy_mean}(c)), such seeming variation can in turn help to reach plausible and consistent results by averaging over different colors (\fref{fig:toy_mean}(b)).

Our query-based colorization can be conducted in a simple way. After sampling each batch of rays corresponding to image patch $\cP(\bu, s)$, we first render the monochromatic image patch $\hat{\bL}_\cP$ based on the luminance and density representation. Then we colorize this monochromatic image patch by feeding it into the colorization module as follows:
\begin{equation}
    \mathbf{B}_\cP = \cF(\hat{\bL}_\cP),
    \label{eq:onlinequery}
\end{equation}
where $\cF(\cdot)$ denotes the colorization module and $\mathbf{B}_\cP$ denotes the colorized image patch.

\begin{figure}[t]
    \centering

    \includegraphics[width=1.\linewidth]
    {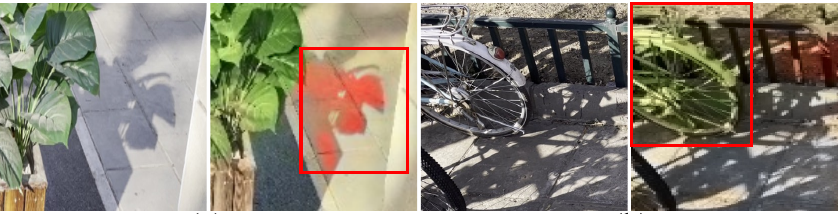}
    \caption{
    Outliers from colorized image patches. For both examples, the left is the reference image and the right is the colorized image with outliers, marked by red rectangles.
    }
    \label{fig:toy_outlier}
\end{figure}

\paragraph{Histogram-guided purification. }
The query-based colorization has been able to produce plausible colorized samples across different views. However, due to incorrect understanding of the scene, the colorization module sometimes yield outliers with different layout or illumination that may undermine the colorization process, demonstrated in \fref{fig:toy_outlier}.

We propose to purify such outliers by histogram similarity comparison. Before training, we sample patches with $s=0.7$ and generate base color images $\mathbf{b}$. With large perceptual field, $\mathbf{b}$ capture full semantics of the scene, so few outliers occur. In the training epochs, we sample $\cP$ with $s\in[0.3, 0.7]$ to colorize the details. After acquiring the colorized patches $\mathbf{B}_\cP$, we calculate the histogram similarity between $\mathbf{B}_\cP$ and $\mathbf{b}$:
\begin{equation}
    \label{eq:histcomparison}
    d(\mathbf{B}_\cP) = \frac{\sum_j (\Delta_H(\mathbf{b})\odot\Delta_H(\mathbf{B}_\cP))}{\sqrt{ \sum_j (\Delta_H(\mathbf{b}))^2 \cdot \sum_j (\Delta_H(\mathbf{B}_\cP))^2  }},
\end{equation}
where $\odot$ denotes element-wise multiplication and $\Delta_H(\cdot)$ is the normalized color histogram of a given image.
$j$ is histogram bin index.

As demonstrated in \fref{fig:pipeline}, by comparing $\Delta_H(\mathbf{B}_\cP)$ with $\Delta_H(\mathbf{b})$, the purification module can exclude outliers deviating from samples based on the selection scheme as follows: 
\begin{equation}
\label{eq:bigh module}
\mathcal{H}(\mathbf{B}_\cP) = 
\left\{
\begin{array}{rc}
1,    & \mathop{\max}(d(\mathbf{B}_\cP)) > T, \\
0,    & \mathrm{otherwise}.
\end{array}
\right.
\end{equation}
We empirically choose $5$ base images with $T = 0.80$ in our experiments, $\mathop{\max}(d(\mathbf{B}_\cP))$ denotes the highest similarity score with the base images. The histogram-guided purification is formulated as $\bY_\cP = \mathcal{H}(\mathbf{B}_\cP)\odot\mathbf{B}_\cP$, 
where $\bY_\cP$ denotes the purified color image patches. With ample plausible color information, we further propose a color injection module, aiming to produce vivid results by properly injecting $\bY_\cP$ to the color representation $\Theta_{\mathrm{Z}}$.

\paragraph{Classification-based color injection. } 
The color injection can be achieved by minimizing the photometric loss similar to \Eref{eq:photo}, which transfers color information from the colorization module to the predicted representation by measuring the differences in their output. Despite the effectiveness of photometric loss in reconstructing radiance fields, it is incapable of producing vivid color~\cite{zhang2016colorful}, since the extensive color distribution inevitably collapses into its mean value during the computation of photometric loss, which renders grayish and less vivid samples from predicted representations.

We propose to preserve extensive color distribution by considering color injection as a classification task. To fit the classification objective, we first quantize the possible $ab$ space with grid size 10 and keep $Q=313$ colors which are in-gamut, denoted as $\omega_q\in \mathbb{R} ^ 2$, where $q \in \{1, 2, \ldots, Q\}$ is the index of quantized $ab$ candidates. We change the output channel of the color representation $\Theta_{Z}(\cdot)$ to $Q$ channels as probability scores of the possible color labels. For each sampling patch $\cP$, we predict a probability distribution of quantized $ab$ colors, denoted as $\hat{\bZ}_\cP \in [0, 1]^{K\times K\times Q}$. 

To supervise $\hat{\bZ}_\cP$ with $\bY_\cP$, for each pixel $\mathbf{p}$, we find $5$ quantized colors closest to $\bY_\cP(\mathbf{p})$ using the nearest neighbor algorithm, and use their distances as weights to generate the soft label $\bZ_\cP(\mathbf{p})$, such soft-label operation is denoted as $\cS(\cdot)$, \ie, $\bZ_\cP(\mathbf{p}) = \cS(\bY_\cP(\mathbf{p}))$.

We formulate the color classification loss as follows:
\begin{equation}
\label{eq: color classification loss}
\cL_{\mathrm{cl}} = -\!\!\!\sum_{\mathbf{p} \in \cP, q}\!\!(\log(\hat{\bZ}_\cP^q(\mathbf{p})) - \log(\bZ_\cP^q(\mathbf{p}))\bZ_\cP^q(\mathbf{p}).
\end{equation}
In the inference stage, we simply choose the color with the largest possibility score from $Q$ candidates, and take the $ab$ values of that color as our prediction:
\begin{equation}
\label{eq: z to y}
    \hat{\bY}_\cP(\mathbf{p}) = \mathbf{\omega}_q ,\mathrm{where}\,\, q = \mathop{\arg\max}\limits_{q}\, \hat{\bZ}_\cP^q(\mathbf{p}), 
\end{equation}
and $\bZ_\cP^q(\mathbf{p})$ is the probability score for $\omega_{q}$ color. We formulate the final results in RGB channel (denoted as $\hat{\bC}_\cP$) by concatenating $\hat{\bY}_\cP$ with $\hat{\bL}_\cP$ and converting the output from $Lab$ to RGB color space: $\hat{\bC}_\cP = \mathrm{Lab2RGB}(\mathrm{concat}\{\hat{\bL}_\cP, \hat{\bY}_\cP\}).$

\subsection{Implementation details}
\label{subsec: implementation details}
We implement our pipeline using PyTorch. We integrate the colorization modules based on their released implementation, and freeze their model weights along training. Following the design in NeRF~\cite{MildenhallSTBRN20}, an eight-layer MLP with 256 channels is used for points encoding, and the luminance and color MLPs have two layers with 128 channels for directional encoding. 
Along each ray, we sample 64 points to train a ``coarse'' network and 64 additional importance sampling points to train a ``fine'' network. An image patch with $K=128$ size is sampled in a batch. Positional encoding is applied to input location and direction similar to NeRF~\cite{MildenhallSTBRN20}. 
We optimize our model for 30 epochs on one NVIDIA TITAN RTX GPU.

\begin{table*}[t]
\begin{center}

\caption{Quantitative comparison results on synthetic monochromatic data. $\uparrow$ ($\downarrow$) means higher (lower) is better. The best performances are highlighted in \textbf{bold}.}

\label{tab:quan}
\begin{tabular}{c|c|c|c|c|c|c}
\hline
    Category & Method     & PSNR$\uparrow$ & SSIM$\uparrow$ & LPIPS$\downarrow$ & Colorful$\uparrow$ & $\Delta$ Colorful$\downarrow$  \\ \hline

  Comparison  &   Vid~\cite{Lei2019videocolor}+NeRF  &   17.78   &  0.63     &  0.31     & 12.88  &  23.59     \\ 
  Comparison  &  ARF~\cite{zhang2022arf}             &   17.82   &  0.54     &  0.37     & 36.33  &   N/A      \\ 
  Comparison  &  CLIP-NeRF~\cite{wang2022clipnerf}   &   17.76   &  0.76     &  0.31     & 30.92  &   12.18     \\ 
  Comparison  &   \CT~\cite{WengCT2}+NeRF      &   18.90   &  0.80    &  0.32     & 48.52  &   18.24   \\ \hline
Ablation    &   w/o  histogram-guided purification    &   17.74   &   0.77     &  0.25 & 54.41  & 19.20     \\ 
Ablation    &   w/o  classification-based color injection    &  19.22   &  \textbf{0.81}     &  0.22     & 49.28  &  12.82    \\ \hline
Ours    &   ColorNeRF            &   \textbf{20.76}   &  \textbf{0.81}     &  \textbf{0.21}     & \textbf{55.40}  &   \textbf{12.15}     \\ \hline
\end{tabular}
\end{center}
\end{table*}

\section{Experiments}
\label{subsec: experiment settings}

\begin{figure}[t]
\centering
\includegraphics[width=\linewidth]{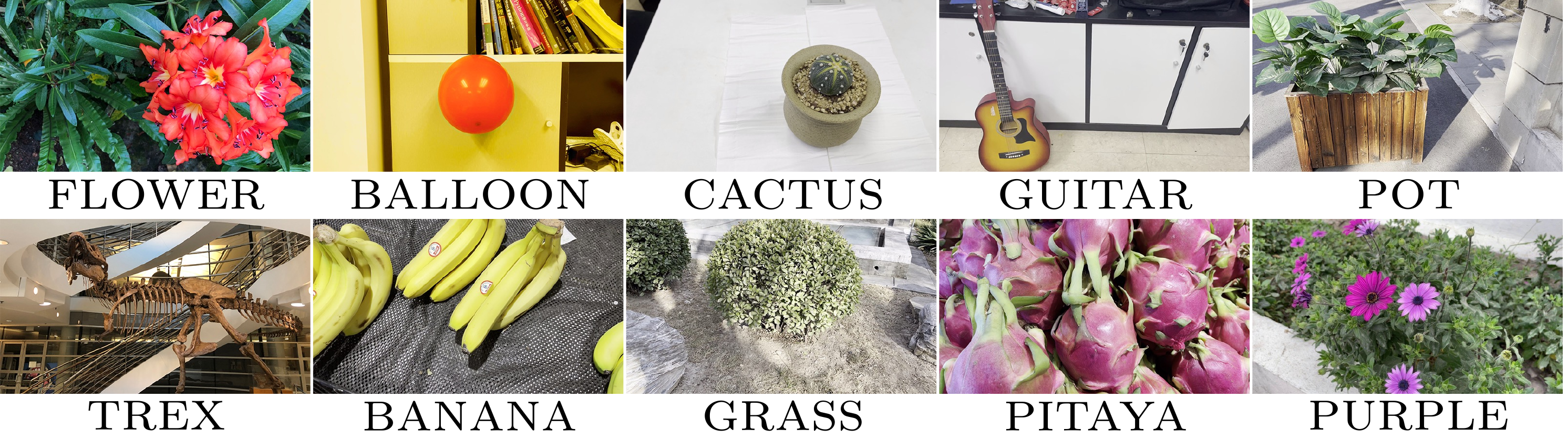}
\caption{One image sample for each scene in our dataset. The names of the scenes are listed below the image.}\label{fig:dataset}
\end{figure}

\paragraph{Dataset.} To conduct quantitative evaluation, we first employ \textbf{synthetic monochromatic data}. Two samples from the conventional LLFF dataset~\cite{mildenhall2019llff} are employed (\textsc{flower} and \textsc{trex}). We additionally capture 8 scenes by following the instructions of LLFF~\cite{mildenhall2019llff}, and each scene consists of $30$ to $50$ viewpoints. A sample of each scene can be found in \cref{fig:dataset}. During the experiments, the samples in the above scenes are transformed into their monochromatic counterparts, and their original colorful version is used as the ground truth. 
To evaluate whether ColorNeRF is effective for \textbf{real monochromatic data} (without ground truth color) directly produced by imaging devices. We further capture a scene using the spike camera, a novel type of neuromorphic sensor recording scene radiance as colorless neural spikes, which can be integrated to monochromatic images~\cite{huang20221000}. In addition, two multi-view scenes collected from old movies\footnote{``Breathless'' by Jean-Luc Godard, 1960 and ``The Man Who Sleeps" by Georges Perec, 1974} are employed to demonstrate our potential in rejuvenating old digital archives. 
All scenes are first processed by COLMAP~\cite{schoenberger2016sfm, schoenberger2016mvs} for pose estimation. Synthetic data are used for quantitative and qualitative evaluations.

\begin{figure}[t]
    \centering
    \includegraphics[width=1\linewidth]{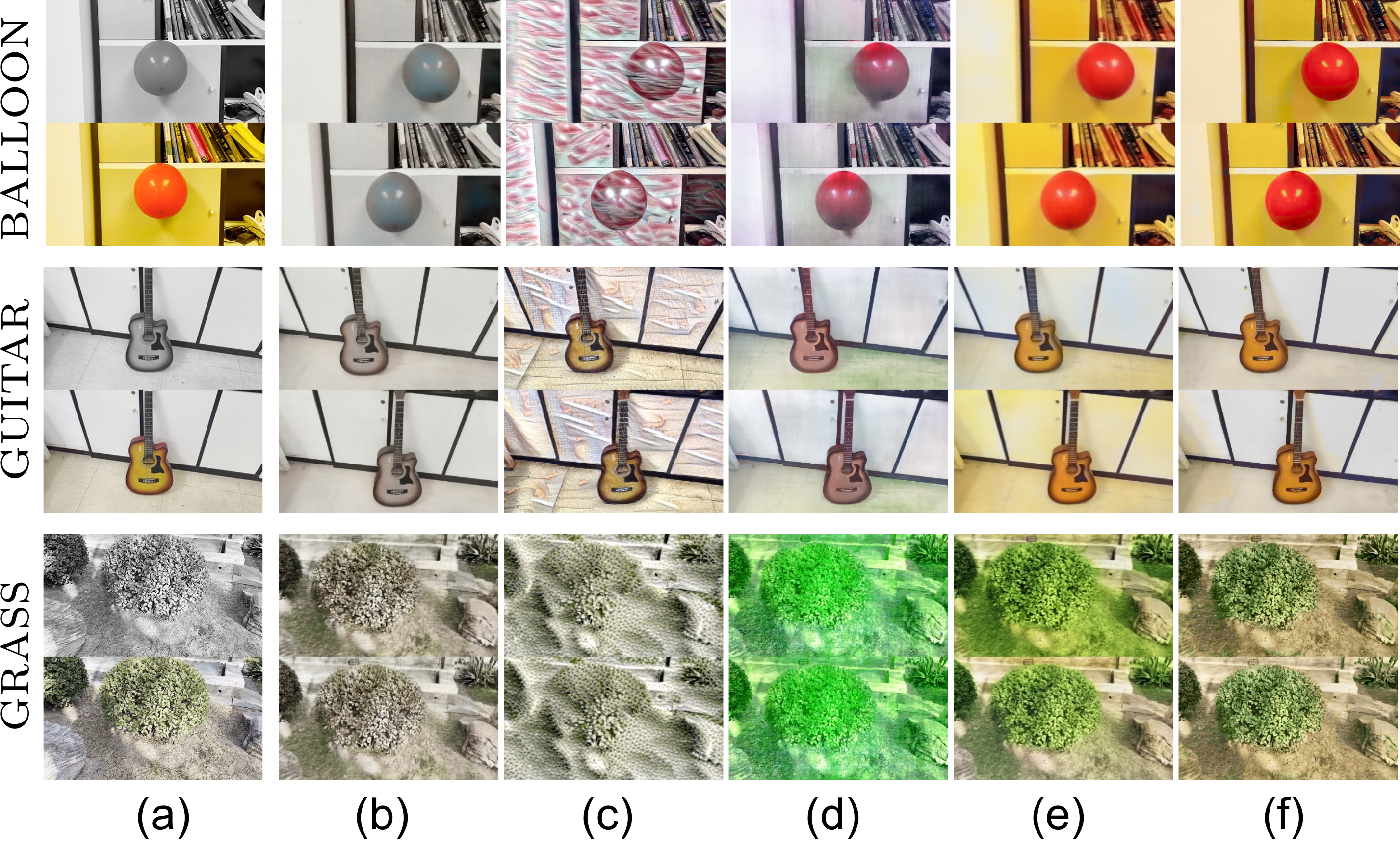}
    \caption{Qualitative comparison with selected baselines. In each scene, (a) is the monochormatic input and reference image (not used in training); (b-f) show two novel synthesised views of the compared methods. (b): Vid~\cite{Lei2019videocolor} + NeRF; (c): ARF~\cite{zhang2022arf}; (d): CLIP-NeRF~\cite{wang2022clipnerf}; (e): \CT~\cite{WengCT2}+NeRF; (f) ColorNeRF (ours).
    }
    \label{fig:qual}
\end{figure}

\paragraph{Baselines.} We compare ColorNeRF against the following baselines: 1) \textbf{CLIP-NeRF}~\cite{wang2022clipnerf}, a 3D object manipulation method with color editing ability; 2) \textbf{ARF}~\cite{zhang2022arf}, a style transfer NeRF method using a ground truth image as the reference style image; 
3) \textbf{\CT~\cite{WengCT2}+NeRF}, results using the ``colorize-then-fuse" paradigm with the same colorization model \CT~\cite{WengCT2} used in our pipeline; 4) \textbf{Vid~\cite{Lei2019videocolor}+NeRF}, results using the ``colorize-then-fuse" paradigm with \sota{} automatic video colorization work~\cite{Lei2019videocolor}. 
We do not compare with palette-based color editing NeRFs~\cite{tojo2022posternerf}, since their palette extraction module~\cite{TanEG18, TanLG17} could not extract palette from monochromatic images. 
\paragraph{Error metrics. } We measure the performance of our colorized results following the conventionally used metrics in colorization and implicit radiance fields.  
PSNR~\cite{huynh2008scope}, SSIM~\cite{DBLP:journals/tip/WangBSS04} and LPIPS~\cite{DBLP:conf/cvpr/ZhangIESW18} are used to measure the image quality of the results; 
Colorful Score~\cite{DBLP:conf/hvei/HaslerS03} reflects the vividness of the colorized images. The absolute colorfulness score difference ($\Delta$ Colorful) of the ground truth images and predicted ones could also show how close predictions are to ground truth in terms of vividness.

While pixel-level metrics, such as PSNR and SSIM, are commonly utilized for quantitative evaluation, it has been recognized that these metrics may not accurately reflect the true performance of colorization techniques~\cite{messaoud2018structural, su2020instance,wu2021towards}. Hence, in order to validate the performance of the compared methods, a user study is conducted.

\paragraph{Quantitative experiments. }
The quantitative results on synthetic monochromatic data are reported in \Tref{tab:quan}. Our model achieves better performance in all metrics. \CT~\cite{WengCT2}+NeRF has the second-best performance, but their major drawbacks lie in the inconsistency observed in \fref{fig:qual}. 
ARF~\cite{zhang2022arf} utilizes a ground truth image as the style image, hence it is unsuitable to compare with other methods on colorfulness metrics. 

\paragraph{Novel view synthesis.}
In \fref{fig:qual}, we show the novel view synthesis results on our model and the compared baselines. It is clear that our model could yield the most plausible and vivid colorization results. In Vid~\cite{Lei2019videocolor}+NeRF, the video colorization module fails to colorize vivid results, probably due to it is over-fitted on its training dataset and has a major domain gap with real-world custom scenes. ARF~\cite{zhang2022arf} focuses mainly on extracting the pattern in the style image, it could get artistic results, but the results are not plausible since the geometry of the scenes are influenced by the style patterns. CLIP-NeRF~\cite{wang2022clipnerf} could not extract color information from monochromatic images. Hence it yields less vivid results. In \CT~\cite{WengCT2}+NeRF, the results are not plausible since the color is flickering when the view direction changes. 
We refer the readers to the project page for more comprehensive results in our dataset with higher resolution and more synthesised views.

\label{subsec:ablation study}
\begin{figure}[t]
\centering
\includegraphics[width=1\linewidth]{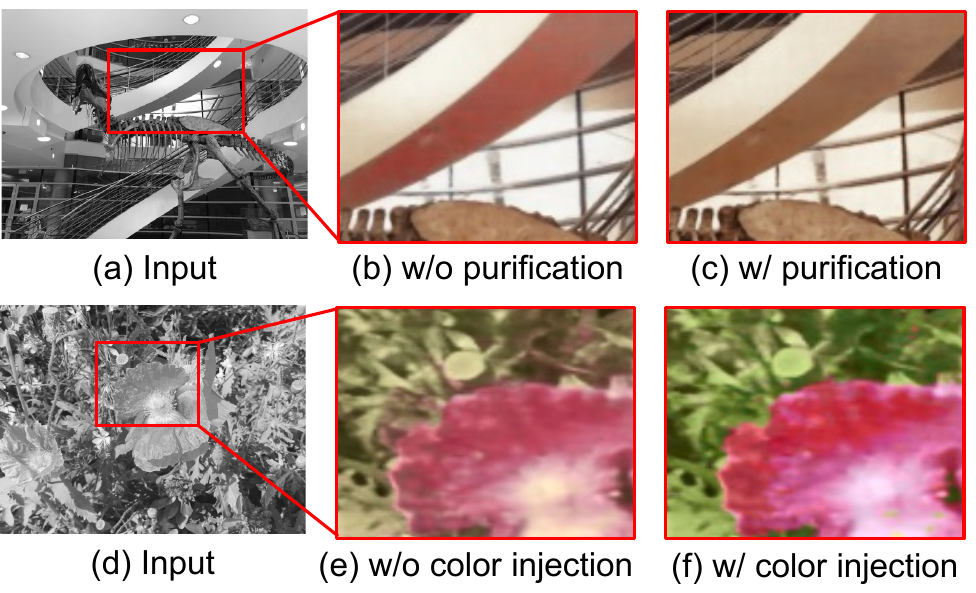}
\caption{Ablation study. Comparison between (b) and (c) shows the histogram-guided purification module could purify undesired color; comparison between (e) and (f) exhibits the classification-based color injection module produces more vivid outputs.}
\label{fig:ab}
\end{figure}

\begin{table}[t]
\caption{User study results. Higher score means better performance, and the best scores are highlighted in \textbf{bold}. Our model exhibits superior performance in terms of plausibility and vividness compared to other methods. (a): Vid~\cite{Lei2019videocolor}+NeRF; (b): ARF~\cite{zhang2022arf}; (c): CLIP-NeRF~\cite{wang2022clipnerf}; (d): \CT~\cite{WengCT2}+NeRF; (e): ColorNeRF (ours).}
\label{tab:user study}
\begin{center}
\begin{tabular}{c|c|c|c|c|c}
\hline
Method   & (a) & (b) & (c) & (d) & (e) \\ \hline
Plausibility & 1.70 & 2.23 & 3.10 & 3.25 & \textbf{4.26}  \\ 
Vividness    & 3.00 & 2.35 & 2.55 & 3.40 & \textbf{4.67} \\ \hline
\end{tabular}
\end{center}

\end{table}

\paragraph{Ablation study. }
The quantitative and qualitative ablation results are presented in \Tref{tab:quan} and \fref{fig:ab}, respectively. The absence of the histogram-based purification module in quantitative experiments results in a high Colorful score, but the $\Delta$ Colorful score is also high, indicating the presence of undesired colors, such as the red area in \fref{fig:ab}(b). On the other hand, the omission of the classification-based color injection in the ablation experiments leads to a decrease in the Colorful score, indicating a less vivid performance, \eg, the yellowish leaves in \fref{fig:ab}(e).

\paragraph{User study. }
In addition to quantitative and qualitative comparisons, we conduct user study experiments to assess whether our results are preferred by human observers. The experiment set is composed of the 10 scenes in synthetic monochromatic data. For each scene, we provide 3 synthesised novel views colorized by 5 different methods: Vid~\cite{Lei2019videocolor}+NeRF, ARF~\cite{zhang2022arf}, CLIP-NeRF~\cite{wang2022clipnerf}, \CT~\cite{WengCT2}+NeRF and Ours. Participants are asked to score 1-5 (higher means better performance) on the results in terms of plausibility and vividness. The order of displayed methods is shuffled in each scene. Each experiment is completed by 50 participants. Results in \Tref{tab:user study} show our method outperforms other methods in terms of plausibility and vividness.


\begin{figure}[t]
    \centering
    \includegraphics[width=1\linewidth]{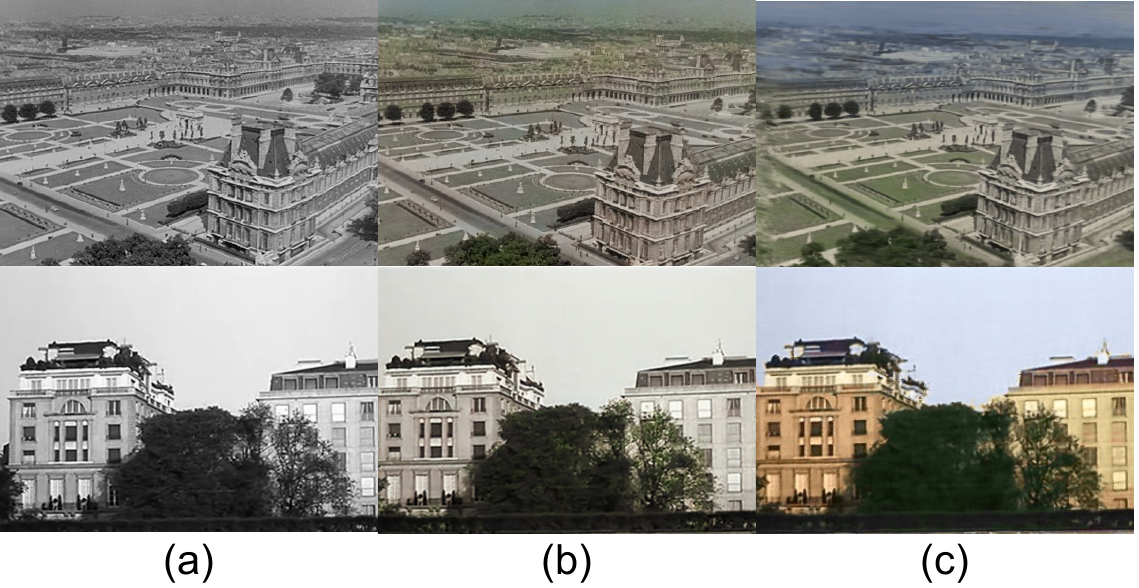}
    \caption{Qualitative comparison with Vid~\cite{Lei2019videocolor} + NeRF on two old movie clips. (a): input; (b): Vid~\cite{Lei2019videocolor} + NeRF; (c): ColorNeRF (ours). 
    Please refer to the project page for animation results.
    }
    \label{fig:app}
\end{figure}

\begin{figure}[t]
    \centering
    \includegraphics[width=0.32\linewidth]{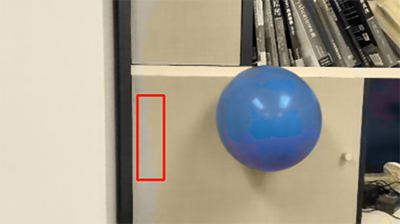}
        \includegraphics[width=0.32\linewidth]{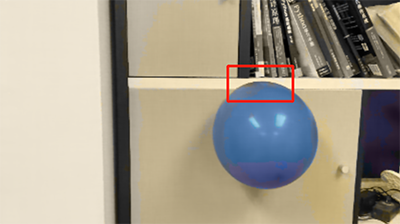}
        \includegraphics[width=0.32\linewidth]{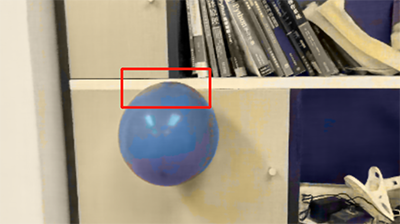}
    \caption{Three novel views of ColorNeRF using color information from L-CoDer~\cite{chang2022lcoder} on {\sc balloon} scene are shown. The text guidance used in L-CoDer is ``Blue balloon on white wall".}

    \label{fig:limitation}
\end{figure}

\paragraph{Results on real data. }
We show the results of our model on real monochromatic data in \fref{fig:app}, which also demonstrate our model's applications on creating colors for rejuvenating old digital archives in the form of radiance fields.

\section{Conclusions}
\label{sec: conclusion}
In this paper, we introduce ColorNeRF, a novel approach capable of generating plausible and vivid radiance fields from monochromatic images. Our approach employs a representation prediction framework, incorporating a query-based colorization module, a histogram-guided purification module, and a classification-based color injection module, to ensure the plausibility and vividness of the results. Extensive experiments are conducted to validate the advantages and broad applications of our model.

\paragraph{Limitations and future work.}
Theoretically, an arbitrary colorization network could be incorporated with ColorNeRF. In \fref{fig:limitation}, we show the novel views generated by ColorNeRF using language-guided colorization model L-CoDer~\cite{chang2022lcoder}. 
Although our model produces plausible and vivid outcomes, there are some undesired artifacts in the results. This is resulted from the inferior performance of L-CoDer~\cite{chang2022lcoder} under cases with uncertain background colors, which also demonstrates that the effectiveness of ColorNeRF is contingent upon the performance of the external colorization module. The advancement of 2D colorization models is expected to mitigate this issue by generating more plausible results.

\section*{Acknowledgements} This work is supported by the National Natural Science Foundation of China under Grand No. 62088102 and 62136001. Renjie Wan is supported by the National Natural Science Foundation of China under Grant No. 62302415, Guangdong Basic and Applied Basic Research Foundation under Grant No. 2022A1515110692, and the Blue Sky Research Fund of HKBU under Grant No. BSRF/21- 22/16. Yakun Chang is supported by the National Natural Science Foundation of China under Grant No. 62301009.


\bibliography{CameraReady/LaTeX/aaai24}

\begin{thebibliography}{48}
\providecommand{\natexlab}[1]{#1}

\bibitem[{Anwar et~al.(2020)Anwar, Tahir, Li, Mian, Khan, and
  Muzaffar}]{anwar2020image}
Anwar, S.; Tahir, M.; Li, C.; Mian, A.; Khan, F.~S.; and Muzaffar, A.~W. 2020.
\newblock Image colorization: A survey and dataset.
\newblock \emph{arXiv preprint arXiv:2008.10774}.

\bibitem[{Cao et~al.(2017)Cao, Zhou, Zhang, and Yu}]{cao2017unsupervised}
Cao, Y.; Zhou, Z.; Zhang, W.; and Yu, Y. 2017.
\newblock Unsupervised diverse colorization via generative adversarial
  networks.
\newblock In \emph{Machine Learning and Knowledge Discovery in Databases:
  European Conference}.

\bibitem[{Chang et~al.(2022)Chang, Weng, Li, Li, and Shi}]{chang2022lcoder}
Chang, Z.; Weng, S.; Li, Y.; Li, S.; and Shi, B. 2022.
\newblock {L-CoDer}: Language-based colorization with color-object decoupling
  transformer.
\newblock In \emph{Proc. of European Conference on Computer Vision}.

\bibitem[{Chen et~al.(2018)Chen, Shen, Gao, Liu, and Liu}]{chen2018language}
Chen, J.; Shen, Y.; Gao, J.; Liu, J.; and Liu, X. 2018.
\newblock Language-based image editing with recurrent attentive models.
\newblock In \emph{Proc. of Computer Vision and Pattern Recognition}.

\bibitem[{Cheng, Yang, and Sheng(2015)}]{DBLP:conf/iccv/ChengYS15}
Cheng, Z.; Yang, Q.; and Sheng, B. 2015.
\newblock Deep colorization.
\newblock In \emph{Proc. of International Conference on Computer Vision}.

\bibitem[{Fan et~al.(2022)Fan, Wang, Jiang, Gong, Xu, and
  Wang}]{fan2022nerfsos}
Fan, Z.; Wang, P.; Jiang, Y.; Gong, X.; Xu, D.; and Wang, Z. 2022.
\newblock {NeRF-SOS}: Any-view self-supervised object segmentation on complex
  scenes.
\newblock \emph{CoRR}, abs/2209.08776.

\bibitem[{Geonung et~al.(2022)Geonung, Kyoungkook, Seongtae, Hwayoon, Sehoon,
  Jonghyun, Seung-Hwan, and Sunghyun}]{Kim2022BigColor}
Geonung, K.; Kyoungkook, K.; Seongtae, K.; Hwayoon, L.; Sehoon, K.; Jonghyun,
  K.; Seung-Hwan, B.; and Sunghyun, C. 2022.
\newblock {BigColor}: Colorization using a generative color prior for natural
  images.
\newblock In \emph{Proc. of European Conference on Computer Vision}.

\bibitem[{Gong et~al.(2023)Gong, Wang, Han, and Dou}]{Gong2023RecolorNeRFLD}
Gong, B.; Wang, Y.; Han, X.; and Dou, Q. 2023.
\newblock {RecolorNeRF}: Layer decomposed radiance fields for efficient color
  editing of {3D} scenes.
\newblock \emph{Proc. of ACM International Conference on Multimedia}.

\bibitem[{Guadarrama et~al.(2017)Guadarrama, Dahl, Bieber, Shlens, Norouzi, and
  Murphy}]{guadarrama2017pixcolor}
Guadarrama, S.; Dahl, R.; Bieber, D.; Shlens, J.; Norouzi, M.; and Murphy, K.
  2017.
\newblock {PixColor}: Pixel recursive colorization.
\newblock In \emph{Proc. of British Machine Vision Conference}.

\bibitem[{Hasler and S{\"{u}}sstrunk(2003)}]{DBLP:conf/hvei/HaslerS03}
Hasler, D.; and S{\"{u}}sstrunk, S. 2003.
\newblock Measuring colorfulness in natural images.
\newblock In \emph{Human Vision and Electronic Imaging VIII}.

\bibitem[{Huang et~al.(2022)Huang, Zheng, Yu, Chen, Li, Xiong, Ma, Zhao, Dong,
  Zhu et~al.}]{huang20221000}
Huang, T.; Zheng, Y.; Yu, Z.; Chen, R.; Li, Y.; Xiong, R.; Ma, L.; Zhao, J.;
  Dong, S.; Zhu, L.; et~al. 2022.
\newblock 1000$\times$ faster camera and machine vision with ordinary devices.
\newblock \emph{Engineering}.

\bibitem[{Huang, Zhao, and Liao(2022)}]{huang2022unicolor}
Huang, Z.; Zhao, N.; and Liao, J. 2022.
\newblock Unicolor: A unified framework for multi-modal colorization with
  transformer.
\newblock \emph{ACM Transactions on Graphics}.

\bibitem[{Huynh-Thu and Ghanbari(2008)}]{huynh2008scope}
Huynh-Thu, Q.; and Ghanbari, M. 2008.
\newblock Scope of validity of {PSNR} in image/video quality assessment.
\newblock \emph{Electronics letters}.

\bibitem[{Iizuka, Simo{-}Serra, and Ishikawa(2016)}]{iizuka2016let}
Iizuka, S.; Simo{-}Serra, E.; and Ishikawa, H. 2016.
\newblock {Let there be color}!: Joint end-to-end learning of global and local
  image priors for automatic image colorization with simultaneous
  classification.
\newblock \emph{ACM Transactions on Graphics}.

\bibitem[{Ironi, Cohen-Or, and Lischinski(2005)}]{ironi2005colorization}
Ironi, R.; Cohen-Or, D.; and Lischinski, D. 2005.
\newblock Colorization by example.
\newblock \emph{Rendering techniques}.

\bibitem[{Kobayashi, Matsumoto, and
  Sitzmann(2022)}]{kobayashi2022distilledfeaturefields}
Kobayashi, S.; Matsumoto, E.; and Sitzmann, V. 2022.
\newblock Decomposing {NeRF} for editing via feature field distillation.
\newblock In \emph{Proc. of Neural Information Processing Systems}.

\bibitem[{Kuang et~al.(2022)Kuang, Luan, Bi, Shu, Wetzstein, and
  Sunkavalli}]{Kuang2022PaletteNeRFPA}
Kuang, Z.; Luan, F.; Bi, S.; Shu, Z.; Wetzstein, G.; and Sunkavalli, K. 2022.
\newblock {PaletteNeRF}: Palette-based appearance editing of neural radiance
  fields.
\newblock \emph{Proc. of Computer Vision and Pattern Recognition}.

\bibitem[{Larsson, Maire, and Shakhnarovich(2016)}]{larsson2016learning}
Larsson, G.; Maire, M.; and Shakhnarovich, G. 2016.
\newblock Learning representations for automatic colorization.
\newblock In \emph{Proc. of European Conference on Computer Vision}.

\bibitem[{Lei and Chen(2019)}]{Lei2019videocolor}
Lei, C.; and Chen, Q. 2019.
\newblock Fully automatic video colorization with self-regularization and
  diversity.
\newblock In \emph{Proc. of Computer Vision and Pattern Recognition}.

\bibitem[{Levin, Lischinski, and Weiss(2004)}]{levin2004colorization}
Levin, A.; Lischinski, D.; and Weiss, Y. 2004.
\newblock Colorization using optimization.
\newblock \emph{Proc. of ACM SIGGRAPH}.

\bibitem[{Liu et~al.(2021)Liu, Zhang, Zhang, Zhang, Zhu, and
  Russell}]{liu2021editing}
Liu, S.; Zhang, X.; Zhang, Z.; Zhang, R.; Zhu, J.-Y.; and Russell, B. 2021.
\newblock Editing conditional radiance fields.
\newblock arXiv:2105.06466.

\bibitem[{Luan et~al.(2007)Luan, Wen, Cohen-Or, Liang, Xu, and
  Shum}]{luan2007natural}
Luan, Q.; Wen, F.; Cohen-Or, D.; Liang, L.; Xu, Y.-Q.; and Shum, H.-Y. 2007.
\newblock Natural image colorization.
\newblock In \emph{Proceedings of the 18th Eurographics conference on Rendering
  Techniques}.

\bibitem[{Ma et~al.(2022)Ma, Li, Liao, Zhang, Wang, Wang, and
  Sander}]{ma2022deblur}
Ma, L.; Li, X.; Liao, J.; Zhang, Q.; Wang, X.; Wang, J.; and Sander, P.~V.
  2022.
\newblock {Deblur-NeRF}: Neural radiance fields from blurry images.
\newblock In \emph{Proc. of Computer Vision and Pattern Recognition}.

\bibitem[{Martin-Brualla et~al.(2021)Martin-Brualla, Radwan, Sajjadi, Barron,
  Dosovitskiy, and Duckworth}]{martinbrualla2020nerfw}
Martin-Brualla, R.; Radwan, N.; Sajjadi, M. S.~M.; Barron, J.~T.; Dosovitskiy,
  A.; and Duckworth, D. 2021.
\newblock {NeRF in the Wild}: Neural radiance fields for unconstrained photo
  collections.
\newblock In \emph{Proc. of Computer Vision and Pattern Recognition}.

\bibitem[{Messaoud, Forsyth, and Schwing(2018)}]{messaoud2018structural}
Messaoud, S.; Forsyth, D.; and Schwing, A.~G. 2018.
\newblock Structural consistency and controllability for diverse colorization.
\newblock In \emph{Proc. of European Conference on Computer Vision}.

\bibitem[{Mildenhall et~al.(2019)Mildenhall, Srinivasan, Ortiz-Cayon,
  Kalantari, Ramamoorthi, Ng, and Kar}]{mildenhall2019llff}
Mildenhall, B.; Srinivasan, P.~P.; Ortiz-Cayon, R.; Kalantari, N.~K.;
  Ramamoorthi, R.; Ng, R.; and Kar, A. 2019.
\newblock Local light field fusion: Practical view synthesis with prescriptive
  sampling guidelines.
\newblock \emph{ACM Transactions on Graphics}.

\bibitem[{Mildenhall et~al.(2020)Mildenhall, Srinivasan, Tancik, Barron,
  Ramamoorthi, and Ng}]{MildenhallSTBRN20}
Mildenhall, B.; Srinivasan, P.~P.; Tancik, M.; Barron, J.~T.; Ramamoorthi, R.;
  and Ng, R. 2020.
\newblock {NeRF}: Representing scenes as neural radiance fields for view
  synthesis.
\newblock In \emph{Proc. of European Conference on Computer Vision}.

\bibitem[{Niemeyer and Geiger(2021)}]{Niemeyer2021GIRAFFE}
Niemeyer, M.; and Geiger, A. 2021.
\newblock {GIRAFFE}: Representing scenes as compositional generative neural
  feature Fields.
\newblock In \emph{Proc. of Computer Vision and Pattern Recognition}.

\bibitem[{Pearl, Treibitz, and Korman(2022)}]{pearl2022nan}
Pearl, N.; Treibitz, T.; and Korman, S. 2022.
\newblock {NAN}: Noise-aware {NeRFs} for burst-denoising.
\newblock In \emph{Proc. of Computer Vision and Pattern Recognition}.

\bibitem[{Sch\"{o}nberger and Frahm(2016)}]{schoenberger2016sfm}
Sch\"{o}nberger, J.~L.; and Frahm, J.-M. 2016.
\newblock Structure-from-motion revisited.
\newblock In \emph{Proc. of Computer Vision and Pattern Recognition}.

\bibitem[{Sch\"{o}nberger et~al.(2016)Sch\"{o}nberger, Zheng, Pollefeys, and
  Frahm}]{schoenberger2016mvs}
Sch\"{o}nberger, J.~L.; Zheng, E.; Pollefeys, M.; and Frahm, J.-M. 2016.
\newblock Pixelwise view selection for unstructured multi-view stereo.
\newblock In \emph{Proc. of European Conference on Computer Vision}.

\bibitem[{Schwarz et~al.(2020)Schwarz, Liao, Niemeyer, and
  Geiger}]{schwarz2020graf}
Schwarz, K.; Liao, Y.; Niemeyer, M.; and Geiger, A. 2020.
\newblock {GRAF}: Generative radiance fields for {3D}-aware image synthesis.
\newblock In \emph{Proc. of Neural Information Processing Systems}.

\bibitem[{Su, Chu, and Huang(2020)}]{su2020instance}
Su, J.-W.; Chu, H.-K.; and Huang, J.-B. 2020.
\newblock Instance-aware image colorization.
\newblock In \emph{Proc. of Computer Vision and Pattern Recognition}.

\bibitem[{Tan, Echevarria, and Gingold(2018)}]{TanEG18}
Tan, J.; Echevarria, J.~I.; and Gingold, Y.~I. 2018.
\newblock Efficient palette-based decomposition and recoloring of images via
  {RGBXY}-space geometry.
\newblock \emph{ACM Transactions on Graphics}.

\bibitem[{Tan, Lien, and Gingold(2017)}]{TanLG17}
Tan, J.; Lien, J.; and Gingold, Y.~I. 2017.
\newblock Decomposing images into layers via {RGB}-space geometry.
\newblock \emph{ACM Transactions on Graphics}.

\bibitem[{Tojo and Umetani(2022)}]{tojo2022posternerf}
Tojo, K.; and Umetani, N. 2022.
\newblock Recolorable posterization of volumetric radiance fields using
  visibility-weighted palette extraction.
\newblock \emph{Computer Graphics Forum}.

\bibitem[{Vitoria, Raad, and Ballester(2020)}]{vitoria2020chromagan}
Vitoria, P.; Raad, L.; and Ballester, C. 2020.
\newblock {ChromaGAN}: Adversarial picture colorization with semantic class
  distribution.
\newblock In \emph{Proc. of IEEE Winter Conference on Applications of Computer
  Vision}.

\bibitem[{Wang et~al.(2022{\natexlab{a}})Wang, Chai, He, Chen, and
  Liao}]{wang2022clipnerf}
Wang, C.; Chai, M.; He, M.; Chen, D.; and Liao, J. 2022{\natexlab{a}}.
\newblock {CLIP-NeRF}: Text-and-image driven manipulation of neural radiance
  fields.
\newblock In \emph{Proc. of Computer Vision and Pattern Recognition}.

\bibitem[{Wang et~al.(2022{\natexlab{b}})Wang, Jiang, Chai, He, Chen, and
  Liao}]{wang2022nerf}
Wang, C.; Jiang, R.; Chai, M.; He, M.; Chen, D.; and Liao, J.
  2022{\natexlab{b}}.
\newblock {NeRF-Art:} Text-driven neural radiance fields stylization.
\newblock \emph{arXiv preprint arXiv:2212.08070}.

\bibitem[{Wang et~al.(2004)Wang, Bovik, Sheikh, and
  Simoncelli}]{DBLP:journals/tip/WangBSS04}
Wang, Z.; Bovik, A.~C.; Sheikh, H.~R.; and Simoncelli, E.~P. 2004.
\newblock Image quality assessment: from error visibility to structural
  similarity.
\newblock \emph{IEEE Transactions on Image Processing}.

\bibitem[{Weng et~al.(2022)Weng, Sun, Li, Li, and Shi}]{WengCT2}
Weng, S.; Sun, J.; Li, Y.; Li, S.; and Shi, B. 2022.
\newblock {CT}\({}^{\mbox{2}}\): Colorization transformer via color tokens.
\newblock In \emph{Proc. of European Conference on Computer Vision}.

\bibitem[{Wu et~al.(2021)Wu, Wang, Li, Zhang, Zhao, and Shan}]{wu2021towards}
Wu, Y.; Wang, X.; Li, Y.; Zhang, H.; Zhao, X.; and Shan, Y. 2021.
\newblock Towards vivid and diverse image colorization with generative color
  prior.
\newblock In \emph{Proc. of International Conference on Computer Vision}.

\bibitem[{Yun et~al.(2023)Yun, Lee, Park, and Choo}]{Yun_2023_WACV}
Yun, J.; Lee, S.; Park, M.; and Choo, J. 2023.
\newblock {iColoriT}: Towards propagating local hints to the right region in
  interactive colorization by leveraging vision transformer.
\newblock In \emph{Proc. of IEEE Winter Conference on Applications of Computer
  Vision}.

\bibitem[{Zhang et~al.(2022)Zhang, Kolkin, Bi, Luan, Xu, Shechtman, and
  Snavely}]{zhang2022arf}
Zhang, K.; Kolkin, N.; Bi, S.; Luan, F.; Xu, Z.; Shechtman, E.; and Snavely, N.
  2022.
\newblock {ARF}: Artistic radiance fields.
\newblock In \emph{Proc. of European Conference on Computer Vision}.

\bibitem[{Zhang, Isola, and Efros(2016)}]{zhang2016colorful}
Zhang, R.; Isola, P.; and Efros, A.~A. 2016.
\newblock Colorful image colorization.
\newblock In \emph{Proc. of European Conference on Computer Vision}.

\bibitem[{Zhang et~al.(2018)Zhang, Isola, Efros, Shechtman, and
  Wang}]{DBLP:conf/cvpr/ZhangIESW18}
Zhang, R.; Isola, P.; Efros, A.~A.; Shechtman, E.; and Wang, O. 2018.
\newblock The unreasonable effectiveness of deep features as a perceptual
  metric.
\newblock In \emph{Proc. of Computer Vision and Pattern Recognition}.

\bibitem[{Zhang et~al.(2017)Zhang, Zhu, Isola, Geng, Lin, Yu, and
  Efros}]{zhang2017realtime}
Zhang, R.; Zhu, J.-Y.; Isola, P.; Geng, X.; Lin, A.~S.; Yu, T.; and Efros,
  A.~A. 2017.
\newblock Real-time user-guided image colorization with learned deep priors.
\newblock \emph{ACM Transactions on Graphics}.

\bibitem[{Zhao et~al.(2020)Zhao, Han, Shao, and Snoek}]{zhao2020pixelated}
Zhao, J.; Han, J.; Shao, L.; and Snoek, C.~G. 2020.
\newblock Pixelated semantic colorization.
\newblock \emph{International Journal of Computer Vision}.

\end{thebibliography}

\end{document}